\documentclass[sigconf]{acmart}
\AtBeginDocument{%
  }

\copyrightyear{2025} 
\acmYear{2025} 
\setcopyright{none}
\setcctype{by}
\acmConference[CIKM '25] {Proceedings of the 34th ACM International Conference on Information and Knowledge Management}{ November 10--14, 2025}{Seoul, Republic of Korea.}
\acmBooktitle{Proceedings of the 34th ACM International Conference on Information and Knowledge Management (CIKM '25), November 10--14, 2025, Seoul, Republic of Korea}
\acmISBN{979-8-4007-2040-6/2025/11} 
\acmDOI{10.1145/3746252.3761011}

\usepackage{graphicx,tabularx}
\usepackage{multirow}
\usepackage{multicol}
\usepackage{subcaption}
\usepackage{algorithm,algpseudocode}

\usepackage{url}
\usepackage[utf8]{inputenc}
\usepackage{caption}
\usepackage{amsmath}
\usepackage{amsthm}
\usepackage{booktabs}

\usepackage{natbib}
\usepackage[switch]{lineno}
\usepackage{enumerate}
\usepackage{array}

\usepackage{makecell}

\newcommand{\ours}{DoE}

\usepackage{xcolor}
\definecolor{myred}{RGB}{255,90,90}
\definecolor{myblue}{RGB}{90,90,255}
\definecolor{myyellow}{RGB}{235,190,72}
\definecolor{mygray}{RGB}{160,160,160}

\begin{document}

\title{Unplug and Play Language Models: Decomposing Experts in Language Models at Inference Time}

\author{Nakyeong Yang}
\affiliation{%
  \institution{Seoul National University}
  \city{Seoul}
  \country{Korea}
}
\email{yny0506@snu.ac.kr}

\author{Jiwon Moon}
\affiliation{%
  \institution{Seoul National University}
  \city{Seoul}
  \country{Korea}
  }
\email{wldnjs913@snu.ac.kr}

\author{Junseok Kim}
\affiliation{%
  \institution{Seoul National University}
  \city{Seoul}
  \country{Korea}
  }
\email{kim.junseok@snu.ac.kr}

\author{Yunah Jang}
\affiliation{%
  \institution{Seoul National University}
  \city{Seoul}
  \country{Korea}
  }
\email{vn2209@snu.ac.kr}

\author{Kyomin Jung}
\affiliation{%
  \institution{Seoul National University}
  \city{Seoul}
  \country{Korea}
  }
\email{kjung@snu.ac.kr}





\begin{abstract}
Enabled by large-scale text corpora with huge parameters, pre-trained language models operate as multi-task experts using a single model architecture.
However, recent studies have revealed that certain neurons play disproportionately important roles in solving specific tasks, suggesting that task-relevant substructures can be isolated and selectively activated for each task.
Therefore, we introduce \textbf{Decomposition of Experts (\ours)}, a novel framework that dynamically identifies and activates task-specific experts within a language model to reduce inference cost without sacrificing accuracy.
We first define a task expert as a set of parameters that significantly influence the performance of a specific task and propose a four-step \textbf{\textit{unplug-and-play}} process: (1) receiving a user request, (2) identifying the corresponding task expert, (3) performing inference using the expert-localized model, and (4) restoring the original model and waiting for the next task. Using attribution methods and prompt tuning, \ours~isolates task-relevant neurons, minimizing computational overhead while maintaining task performance.
We assume a setting where a language model receives user requests from five widely used natural language understanding benchmarks, processing one task at a time. In this setup, we demonstrate that \ours~achieves up to a $\times 1.73$ inference speed-up with a 65\% pruning rate, without compromising accuracy.
Comparisons with various task expert localization methods reveal that \ours~effectively identifies task experts, while ablation studies validate the importance of its components. Additionally, we analyze the effects of batch size, token count, and layer types on inference speed-up, providing practical insights for adopting \ours. The proposed framework is both practical and scalable, applicable to any transformer-based architecture, offering a robust solution for efficient task-specific inference.

\begin{figure}[h]
\centering
\includegraphics[width=1.0\linewidth]{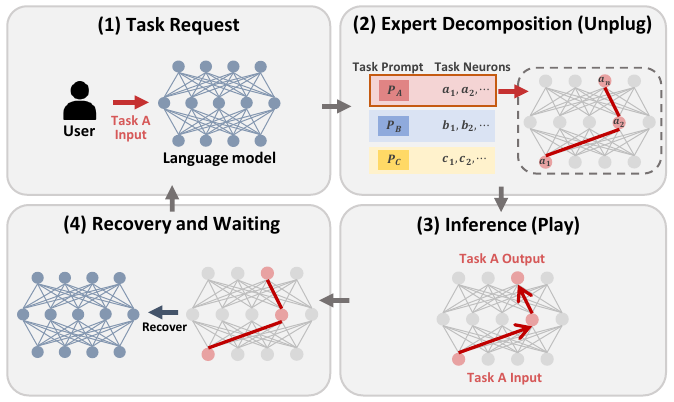}
\vspace{-0.7cm}
\caption{\textbf{Decompostion of Experts (\ours) Framework Overview.} Our framework leverages an unplug-and-play strategy by identifying and activating only the task expert of a language model during the inference procedure when given a user request.
Each color (i.e., \textcolor{myred}{red}, \textcolor{myblue}{blue}, and \textcolor{myyellow}{yellow}) corresponds to a specific task; we utilize a task-specific prompt and neuron indices obtained through a prompt tuning process.
\textcolor{mygray}{Gray-colored} paths in the figure represent task-irrelevant paths that are never used during training and inference, thereby improving inference speed.}
\label{fig:intro1}
\vspace{-0.3cm}
\end{figure}

\end{abstract}

\ccsdesc[500]{Computing methodologies~Natural language processing}

\keywords{Knowledge Localization, Decomposition of Experts, Language Model}




\maketitle

\section{Introduction}

Training on large-scale text corpora with numerous parameters enables recent pre-trained language models to perform a wide range of tasks using a single model architecture \citep{devlin2018bert,liu2019roberta,lewis2019bart,touvron2023llama,openai2023gpt4}.
In this context, previous studies have supposed that each neuron in a language model may have a distinct role; thus, they have tried to investigate the role of individual neurons to explain task-specific knowledge in language models \citep{panigrahi2023task, yang2024mitigating, chen2025identifying}. 
They have addressed various tasks, including sentiment analysis, natural language inference, and social bias detection. Therefore, this line of studies has laid the groundwork for identifying task-relevant neurons in language models by validating that these neurons possess sufficient knowledge to solve a specific task.

Based on these prior studies, we propose a novel framework, \textbf{Decomposition of Experts (\ours)}, that improves the efficiency of language models by dynamically selecting task-specific experts in response to user requests. This approach reduces inference time in single-task scenarios.
We first define a task expert as a subset of parameters that significantly contribute to solving a specific task.
Based on this definition, we describe the DoE framework as an unplug-and-play language model composed of four stages: (1) receiving a user request, (2) identifying the corresponding task expert, (3) performing inference using the expert-localized model, and (4) restoring the original model and waiting for the next task (see Figure~\ref{fig:intro1}).
Specifically, we employ the attribution method \citep{yang2024mitigating} to identify and activate only task-relevant neurons, which we treat as the task expert.
\ours~also adopts prompt tuning \citep{lester2021power} to align task knowledge more effectively within parameters, enabling expert selection using a minimal number of neurons.

While previous pruning approaches have primarily focused on permanently removing unnecessary neurons or connections to reduce the overall parameter count, they typically rely on fixed subnetwork architectures optimized for specific tasks or consistently utilize lightweight models during inference \citep{frankle2018lottery, ma2023llm, chen2024multi}.
In contrast, our method assumes that pre-trained language models inherently possess sufficient multi-task capability. Rather than statically removing parameters, we achieve efficiency by dynamically selecting task-specific expert neurons and temporarily deactivating irrelevant ones (unplug). These neurons can then be reactivated for different tasks, enabling a reversible and task-adaptive computation flow.
From this perspective, our framework offers a novel efficiency strategy that lies at the intersection of pruning and modularization. It is particularly advantageous in multi-task settings where rapid task switching is essential.
\ours~dramatically accelerates the inference speed of language models since it uses only a task expert, which is a small part of the entire model during inference.
For example, to solve a specific task (e.g., sentiment analysis, natural language inference), our method prepends prompt tokens related to the task and activates only a task expert to resolve the task.

Our experimental settings assume that a language model is given user requests about five widely used natural language understanding benchmarks. The experimental results demonstrate that our method significantly enhances inference efficiency without sacrificing the original performance of each task.
Surprisingly, our method improves the inference speed by up to $\times 1.73$ with a 65\% parameter pruning rate while solving a specific task.
In addition, we conduct experiments to compare various task expert selection methods and reveal that the attribution method effectively identifies task experts. 
We also investigate the effects of batch size, the number of tokens, and types of layers on inference speed-up.
Our framework is practical and scalable, as it can be applied to any transformer-based architecture.
In summary, this work makes the following contributions:
\begin{itemize}
\item We first define \textit{task expert} and propose \textit{Decomposition of Expert (\ours)}, a novel language model usage framework that accelerates inference speed by identifying and isolating a task expert within a language model, leveraging an unplug-and-play strategy.
\item We demonstrate that \ours~effectively identifies task experts over other task expert localization methods and enhances the inference speed of each task without compromising task performance.
\item We offer various practical guidelines for adopting \ours~by investigating the effects of hyper-parameters on inference speed-up and evaluating the scalability of the method.
\end{itemize}

\section{Backgrounds}
\subsection{Task Knowledge Localization}
Prior works have shown that language models trained with a vast size of corpora with a huge number of parameters can serve as a multi-task expert \citep{radford2019language, raffel2020exploring}. 
These multi-tasking abilities are attributed to exploiting commonalities and differences recognized while training multiple categories of texts \citep{zhang2022survey, chen2024multi}.
Recent studies have shown that multi-task knowledge is distributed across sparse neurons in language models \citep{liu2023deja, chan2023discoprompt}. Building on this observation, earlier research has investigated the roles of individual parameters in handling specific tasks \citep{panigrahi2023task, chen2025identifying, yang2024mitigating}.

\citet{panigrahi2023task} have suggested a training-based task knowledge localization method, \textit{model grafting}, which identifies task-relevant neurons by training new parameters to mask original parameters.
\citet{chen2025identifying} have investigated the query-relevant neurons by utilizing \textit{intergrated gradients} \citep{sundararajan2017axiomatic}, an explainability method that derives the importance of each neuron when inferring a long text.
\citet{yang2024mitigating} have illuminated bias-relevant neurons by utilizing \textit{attribution} \citep{shrikumar2016not}, an explainability method that derives the importance of each neuron when inferring a specific biased text.
\citet{yang2024mitigating} have verified that the attribution effectively detects bias-relevant neurons and proposed a method applicable to language modeling.

Although \citet{panigrahi2023task, chen2025identifying, yang2024mitigating} have effectively identified task-relevant neurons, they have not proposed practical methods to enhance the usability of language models.
\citet{panigrahi2023task} requires excessive additional storage and computation for newly trained masking parameters, equivalent to the number of model parameters.
In addition, \citet{yang2024mitigating} has a limitation in just eliminating the knowledge-relevant neurons; if we prune a large number of neurons, the performance of a specific task can be significantly degraded.
In other words, this line of studies has only focused on investigating knowledge-relevant neurons, failing to propose a practical method for efficiently utilizing language models during the inference procedure.

\subsection{Language Model Efficiency}
Existing studies have proposed various approaches to enhance the efficiency of language models, which can be categorized into three main groups: (1) knowledge distillation, (2) quantization, and (3) pruning.
Knowledge distillation transfers the knowledge of larger or stronger models into smaller or more efficient ones \citep{gu2023minillm, agarwal2024policy}.
Quantization reduces the precision of model parameters and activations to lower memory and computational costs \citep{shao2023omniquant, yue2024wkvquant, xu2024onebit}.
Pruning eliminates less important parameters to reduce model size and inference time \citep{frankle2018lottery, ma2023llm, chen2024multi}.

While these methods have improved efficiency, they typically rely on statically modifying the model into fixed architectures tailored for specific tasks and deploying task-specific lightweight models consistently at inference time.
In contrast, our approach proposes a new framework of language model usage: \textit{Unplug-and-plug language models}: preserving the full model and dynamically activating only task-relevant neurons, enabling adaptive and reversible task-specific inference without architectural reconfiguration.

\subsection{Parameter-efficient Fine-tuning.}
Parameter-efficient Fine-tuning (PEFT) methods have emerged as a standard of efficient training methods for language models \citep{lester2021power, li2021prefix, liu2021p, hu2021lora, poth2023adapters}.
Among them, prompt tuning methods \citep{lester2021power, li2021prefix, liu2021p} have been adopted for various natural language processing tasks \citep{wang2023fine, bansal2023few}.
Prompt tuning methods only update prompt tokens while maintaining the original parameters of a language model; thus, this characteristic makes the training process memory-efficient and effective in injecting task knowledge into a language model.
However, they are still limited as they are not parameter-efficient during the inference process.
Therefore, existing methods \citep{ma2022xprompt, liang2023prompts} have attempted to accelerate the inference speed of prompt-tuned models by pruning unnecessary prompt tokens.
However, they have only marginally improved inference efficiency, as the number of pruned tokens is a small fraction ($\leq 0.1\%$) of all the parameters of a language model \citep{lester2021power}.
Our \ours~framework adopts the prompt tuning method \citep{lester2021power} to align the original task knowledge when localizing a task expert.

\section{Problem Definition}

\subsection{Unplug-and-play language model}
We define \textit{unplug-and-play language model} as a language model that satisfies the following two conditions: (1) it should perform a single task using only a tiny fraction of its parameters, and (2) the model architecture is fully reversible; thus, it can restore its original parameters after task localization, allowing the model to remain idle to be available for other tasks.
Typical compression methods, such as pruning and knowledge distillation, cannot be considered \textit{unplug-and-play language models} since they statically modify the model into fixed architectures for specific tasks and consistently deploy task-specific lightweight models during inference (irreversible).

\subsection{Prompt Tuning}
We adopt the prompt tuning method to propose a new framework for efficiently utilizing a language model.
Formally, suppose we have a function $\mathcal{P}_{\theta}(y|x)$ that represents a language model, where $\theta$ is the parameters of the language model.
Prompting is the method of adding extra information to the model to condition during its generation of $y$.
Normally, prompting is achieved by adding a series of tokens $P_{\tau} \in \mathbb{R}^{l \times d}$ to the input $x$ as $\mathcal{P}_{\theta}(y|[P_{\tau};x])$, where $l$ and $d$ are token length and feature dimension, and $P_{\tau}$ is trainable parameters for a specific task $\tau$.
In the prompt tuning process, we update only $P_{\tau}$ and get $P'_{\tau}$ to maximize the likelihood of $y$ using gradient descent while freezing the parameters of the language model $\theta$.
After the prompt tuning, we get the new conditional generation function $\mathcal{P}_{\theta}(y|[P'_{\tau};x])$.
However, we assume that there is a minimal parameter combination $\theta_{\tau} \subset \theta$ for solving a specific task $\tau$; thus, we define \textit{task expert} as follows:

\subsection{Task Expert}
Let $\mathcal{P}_{\theta}$ be a language model, where $\theta = \{\theta_{1},...,\theta_{N}\}$ is a set of parameters of it.
Then, $\theta_{\tau} \subset \theta$ is called as a task expert for a task $\tau$ if the below formula is satisfied:

\begin{equation}
\begin{aligned}
    \scriptsize
    \sum_{(x,y) \in \mathcal{D}_{\tau}}\hspace{-0.32cm} \mathcal{L}(\mathcal{P}_{\theta}, x, y) = \epsilon + \hspace{-0.3cm} \sum_{(x,y) \in \mathcal{D}_{\tau}}\hspace{-0.32cm} \mathcal{L}(\mathcal{P}_{\theta_{\tau}}, x, y)
    \small
\end{aligned}
\label{def:taskneuron}
\end{equation}

\noindent where $\mathcal{D}_{\tau}$ denotes the dataset for a task $\tau$ and $\mathcal{L}$ is a score function (e.g., loss function or accuracy) evaluated on the given model. $N$ means the total number of parameters in the language model. $\epsilon$ is a small error value.
In other words, if the scores computed on $\mathcal{P}_{\theta}$ and $\mathcal{P}_{\theta_\tau}$ for the given dataset $\mathcal{D}_{\tau}$ are approximately the same, we can utilize only the task neurons $\theta_{\tau}$ to solve the task $\tau$.

\ours~framework adopts the prompt tuning strategy using only the task expert $\theta_{\tau}$ in a language model; thus, we get the task-localized conditional generation function $\mathcal{P}_{\theta_{\tau}}(y|[P'_{\tau};x])$ instead of $\mathcal{P}_{\theta_{\tau}}(y|x)$.
Notice that since the prompt tuning strategy does not modify the original parameters of language models, a model that adopts the prompt tuning can still be used as a \textit{unplug-and-play language model}.
We can utilize the language model more time-efficiently by using only the sparse neurons within our \ framework.
In addition to being time-efficient, this framework is also memory-efficient since it requires only a small fraction of the original parameters $\theta_{\tau}$ and just a few task-specific prompt tokens $P'_{\tau}$.

\section{Decomposition of Expert (\ours)}
\begin{figure*}[t]
\centering
\includegraphics[width=0.95\textwidth]{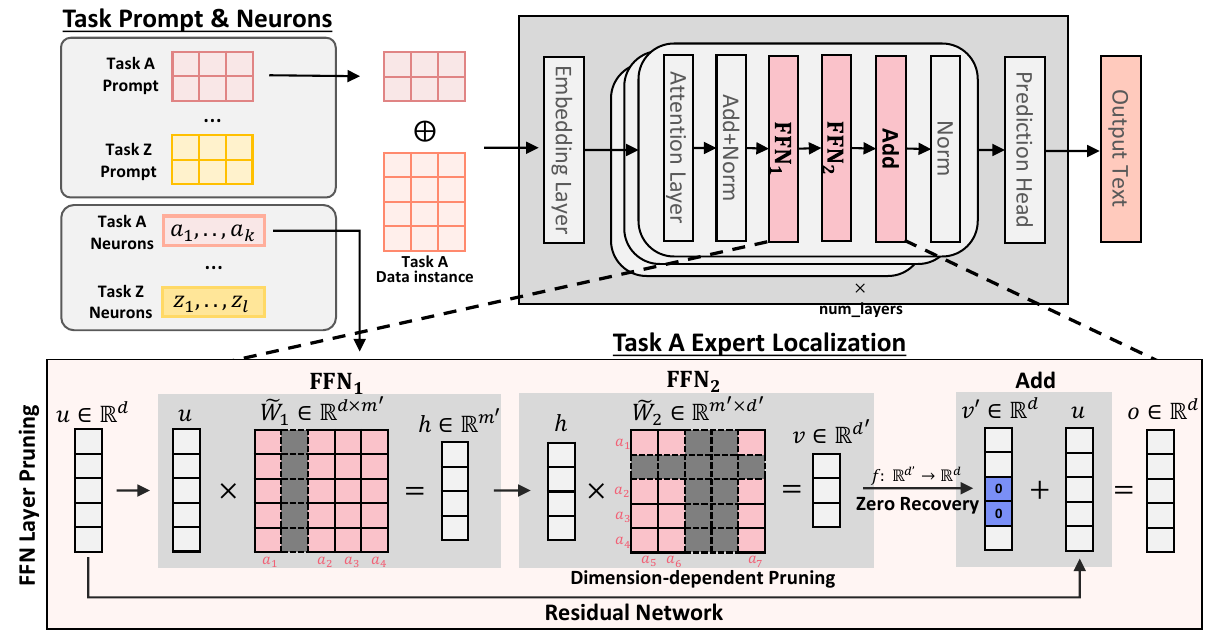}
\caption{\textbf{Task Expert Localization Process of \ours.}
After training a model with the prompt tuning, a task prompt and neurons are assigned to each task, and we utilize only those task-specific resources in the inference procedure. \textcolor{myred}{Red-colored} values in the prune weights, $\tilde{W}_{1}$ and $\tilde{W}_{2}$, mean task neurons for a specific task (task A).
The weight of FFN$_{2}$ is pruned depending on FFN$_{1}$'s dimension, and the output representation of FFN$_{2}$ is zero-recovered (\textcolor{myblue}{Blue-colored} values) to match the output dimension of the residual network (described in Section~\ref{method:details}).
}
\label{fig2}
\end{figure*}

In this section, we describe the process for adopting the \ours~framework.
We first train a language model using the prompt tuning method to align the prior knowledge of a specific task.
After that, we quantify the task relevance of each neuron to determine a task expert in a language model.
Finally, we apply prompt tuning to retrain the prompt tokens, enabling the condensation of task knowledge into the task expert. During this process, only task-relevant neurons are preserved in the original language model.
The method is performed for a set of predefined $T$ tasks; thus, we have $T$ task-specific prompts and $T$ sets of task experts.
After applying the \ours~framework, we can utilize the language model efficiently by using only the task-specific prompt and task expert, accelerating the inference speed for a specific task.
The entire process of \ours~is described in Figure~\ref{fig2} and Algorithm \ref{algorithm:1}.

\subsection{Task Relevance Quantification}
This section describes the process of computing the task relevance of each neuron.
We adopt an attribution method \citep{yang2024mitigating} to quantify the skill relevance of neurons in pre-trained language models.
Formally, given a task-specific prompt tokens $P_{\tau}$ and text input tokens $x$ provided to $\mathcal{P}_{\theta}$, the contribution of the $i$-th neuron to the prediction of an output $y$ is defined as follows:

\begin{equation}
\begin{aligned}
    A^{(P_{\tau},y,x)}_{i}(h)=|h_{i}\times \frac{\partial \mathcal{P}_{\theta}(y|[P_{\tau};x])}{\partial h_{i}}| \\
\end{aligned}
\label{eq:attr_lm}
\end{equation}

\noindent where $h$ corresponds to a hidden representation, and $\partial \mathcal{P}_{\theta}(y|[P_{\tau};x])$ $/\partial h_{i}$ means the gradient of the logit $\mathcal{P}_{\theta}(y|[P_{\tau};x])$ with respect to the $i$-th neuron of the representation $h$.

We adopt transformer variants; thus, activations and gradients are computed for all input token representations.
Therefore, if an input text $x_j$ includes a $K_{j}$ number of tokens, each neuron has a $K_{j}$ number of attributions.
In addition, there are multiple data instances for each task $\tau$; thus, we aggregate attributions for tokens and instances as follows:

\begin{equation}
\begin{aligned}
    A^{(P_{\tau},\mathcal{D}^{\tau})}_{i}(h) = \sum_{j}^{N} \sum_{k}^{K_{j}} \frac{1}{K_j} A^{(P_{\tau},y_{j},x_{j},t_{k})}_{i}(h) \\
\end{aligned}
\label{eq:attr_token_agg}
\end{equation}

\noindent where $A^{(P_{\tau},y_{j},x_{j},t_{k})}_{i}(h)$ means the attribution scores computed for an input text token $t_{k} \in  x_{j}$.
$\mathcal{D}^{\tau}$ and $N$ mean a task-specific dataset and the number of instances in the dataset, respectively.
Therefore, $A^{(P_{\tau},\mathcal{D}^{\tau})}_{i}(h)$ is the attribution score computed for a specific task dataset $\mathcal{D}^{\tau}$.
Although $A^{(P_{\tau},\mathcal{D}^{\tau})}_{i}(h)$ can be computed using the entire dataset, we report the experimental results of computing it using only a significantly small amount of data (e.g., only twenty data samples) to ensure the efficiency of our method.
To simplify the notation, $A^{(P_{\tau},\mathcal{D}^{\tau})}_{i}(h)$ will be abbreviated as $A_{i}$ from this point forward.

\subsection{Task Expert Identification}
After quantifying each neuron's task relevance, we determine task-relevant neurons and eliminate task-irrelevant neurons using a structured pruning method.
We first sort the neurons of the whole target layers by the task relevance scores.
Then, we determine the optimal neuron ratio using the Binary search algorithm to retain as few task neurons as possible without compromising the task knowledge.
Specifically, we compute the accuracy for the validation set $\mathcal{D}^{val}$ to search for the optimal neuron ratio that does not degrade the task performance.
We define a margin $\psi$ and regard $\theta_{\tau}$ as task neurons if the accuracy degradation after neuron localization is smaller than the specified margin $\psi$.

\subsection{Task Expert Localization}
\label{task_neuron_loc}
After identifying task neurons, we eliminate task-irrelevant neurons from a language model using a structured pruning method.
Suppose that a weight matrix $W \in \mathbb{R}^{d \times m}$ is a linear matrix multiplication parameter, and then the matrix after pruning is denoted as $\tilde{W} = (W_{ij})_{\substack{1\leq i\leq d\\ j \in \mathcal{M}}}$, where $\mathcal{M}$ is the set of task-relevant neuron indices about the $W$.
If the bias term $b \in \mathbb{R}^{m}$ is added to the operation for an affine transformation, the bias term can also be pruned by performing the $\tilde{b} = (b_{i})_{i \in \mathcal{M}}$ operation similarly.
The task-localized parameters are used to compute the new representation by performing the transformation operation $h\tilde{W}$ or $h\tilde{W}+\tilde{b}$.
Notice that this method is model-agnostic since all neural network models consist of linear transformation layers.
For example, transformer variants have self-attention and feed-forward network (FFN) modules, all of which include linear matrix multiplication operations. 
After localizing a language model, we further intensify task knowledge of a localized language model using a prompt tuning method.

\subsection{The Entire Process of DoE}
\label{task_neuron_loc_prompt_tuning}
Our \ours~framework follows the three steps: (1) the quantification step of computing task relevance for each neuron; (2) the task expert determination and localization step; (3) the final prompt tuning step for condensing task knowledge. The algorithm for the entire process is described in Algorithm~\ref{algorithm:1}.

\vspace{0.2cm}
\noindent\textbf{(1) The Knowledge Quantification Step.}\hspace{0.2cm}
The attribution method quantifies the contribution of each neuron to the prediction of an output; thus, a language model must have the ability to solve a task in order to quantify the task-specific knowledge of each neuron accurately.
Therefore, we first execute the prompt tuning to align the knowledge in the language model.
Specifically, we conduct the prompt tuning using a frozen language model $\mathcal{P}_{\theta}$ and get an updated task-specific prompt $P_{\tau}'$.
After injecting task knowledge into the language model, we quantify the task relevance $A_{i}$ of each neuron for $\mathcal{P}_{\theta}$ and $P_{\tau}'$ using Equation~\ref{eq:attr_lm} and~\ref{eq:attr_token_agg}.

\vspace{0.2cm}
\noindent\textbf{(2) The Task Expert Localization Step.}\hspace{0.2cm}
After quantifying the task relevance $A_{i}$, we first sort the neurons by the task relevance in descending order.
Then, we select and maintain only the top-$p$ neurons $\theta_{\tau} \subset \theta$ from the language model by pruning task-irrelevant neurons (Section~\ref{task_neuron_loc}).
Note that we use a Binary search algorithm to find the optimal ratio of neurons $p \in [0.0, 0.05, ..., 0.95, 1.0]$.
This search process requires at most five trials, as the pruning ratio is explored in increments of 0.05.

\vspace{0.2cm}
\noindent\textbf{(3) The Knowledge Condensation Step.}\hspace{0.2cm}
During the task expert localization step, we condense the task knowledge to only the task expert.
Although we never update the original parameters of the language model, we can condense the task knowledge into the task expert by updating only $P_{\tau}'$.
We train the task neuron localized model $\mathcal{P}_{\theta_{\tau}}$ and get a new knowledge-condensed task prompt $P_{\tau}''$.
The Binary search algorithm determines the optimal pruning rate after this step.

\algrenewcommand\algorithmicindent{1.5em}%
\algrenewcommand\algorithmicrequire{\textbf{Input:}}
\algrenewcommand\algorithmicensure{\textbf{Output:}}
\begin{algorithm}[t]
 \caption{\ours~Framework}\label{alg:cap}
\begin{algorithmic}[1]
\small 
\Require a language model $\mathcal{P}_{\theta}$; a prompt $P_{\tau}$; a margin $\psi$
\Ensure a task-localized model $\mathcal{P}^{*}_{\theta_{\tau}}$
\vspace{+0.2cm}
\State update $P_{\tau}$ via prompt tuning using $\mathcal{P}_{\theta}$ and get $P'_{\tau}$
\State get the validation accuracy $s_{\tau}$ for $\mathcal{P}_{\theta}$ using $P_{\tau}$
\State compute task relevance $A_{i}$ via Equation~\ref{eq:attr_lm} and \ref{eq:attr_token_agg}
\State sort neurons in descending order by $A_{i}$
\State
\State $l \gets 0$
\State $h \gets 20$
\While {$l \leq r$} \text{\color{blue}// Binary search}
    \State $m \gets \lfloor (l+h) / 2\rfloor$
    \State $p = m \times 0.05$
    \State get top-$p$ task-localized model $\mathcal{P}_{\theta_{\tau}}$
    \State update $P'_{\tau}$ via prompt tuning using $\mathcal{P}_{\theta_{\tau}}$ and get $P''_{\tau}$
    \State get validation accuracy $s'_{\tau}$ for $\mathcal{P}_{\theta_{\tau}}$ using $P''_{\tau}$
    \If {$s_{\tau} - s'_{\tau} \leq \psi$}
        \State $\mathcal{P}_{\theta_{\tau}}^{*} \gets \mathcal{P}_{\theta_{\tau}}$   \text{\color{blue}// The optimal model}
        \State $l \gets m + 1$
    \Else
        \State $h \gets m - 1$
    \EndIf
\EndWhile
\State
\Return $\mathcal{P}_{\theta_{\tau}}^{*}$
\end{algorithmic}
\label{algorithm:1}
\vspace{-0.02cm}
\end{algorithm}




\subsection{Details of the \ours~Framework}
\label{method:details}
\vspace{0.2cm}
\noindent\textbf{Satisfying Dimension Integrity.}\hspace{0.2cm}
This section describes the detailed task expert localization processes; the entire process is shown in Figure~\ref{fig2}.
We prune task-irrelevant neurons from a language model.
Recent language models have been trained using the transformer architecture, but they consist of complex neural computation processes.
For example, sequentially connected networks are dependent on each other in terms of dimensionality.
Suppose there are two sequential feed-forward networks (FFNs), $W_{1} \in \mathbb{R}^{d \times m}$ and $W_{2} \in \mathbb{R}^{m \times d}$.
If we prune the first FFN and get a pruned weight $\tilde{W}_{1} \in \mathbb{R}^{d \times m'}$, this pruning process affects the inconsistency of the dimension with the weight in the second FFN.
Specifically, $W_{2}$ is forced to be pruned as $\tilde{W}_{2} \in \mathbb{R}^{m' \times d}$.
Therefore, we implement the task expert localization algorithm by reflecting the dimension dependency of two FFNs.
In addition, suppose the representation computed by the second FFN, $v \in \mathbb{R}^{d}$, is connected to the representation of a residual connection network, $u \in \mathbb{R}^{d}$.
Then, if we prune the second FFN and get a pruned weight $\tilde{W}_{2} \in \mathbb{R}^{m \times d'}$, this process also triggers the inconsistency of the dimension with the representation transferred by the residual connection, making $v \in \mathbb{R}^{d'}$.
Thus, we insert zero values into the pruned dimension of the second FFN's representation, $v$, to ensure its integrity.
If we do not insert zero values, we can not fully utilize the knowledge of the residual connection.

\vspace{0.2cm}
\noindent\textbf{Accelerating Inference Speed.}\hspace{0.2cm}
Our framework does not just convert the pruned value of weights to zero values but eliminates the entire column (i.e., neuron) of them.
Therefore, our framework accelerates the language model inference speed.
We do not consider other types of networks (e.g., Attention modules, Embedding layer, Language model head) since they play a crucial role in the model's language understanding.
Furthermore, the parameters of FFNs constitute most of the prunable parameters (about $70\%$); thus, pruning even only the FFNs is sufficient.
The experimental results for other modules are described in Section~\ref{sec:module-specific}.

\section{Experiments}

\subsection{Experimental Setup}
\noindent\textbf{Datasets.}\hspace{0.2cm}
We suppose that language models are given requests about five widely-used natural language understanding datasets\footnote{\url{https://huggingface.co/docs/datasets/}} \citep{maas-EtAl:2011:ACL-HLT2011, zhang2015character, wang2018glue} from users. Specifically, we utilize SST-2, IMDB (sentiment analysis); AGNews (topic classification); MRPC (semantic textual matching); CB (natural language inference) to verify the applicability of our method.
We split the train sets and use 10\% of them as a validation set.

\begin{table*}[t]
\centering
\setlength{\tabcolsep}{14pt} 
\resizebox{1.0\textwidth}{!}
{
\begin{tabular}{cc|ccccc|c}
\toprule
 Model &  Method & SST-2 &  IMDB &  AGNews &  MRPC &  CB & \textbf{Avg} \\ \midrule
\multirow{6}{*}{\makecell{BERT}} &  Fine-tuning \small(Single) & 92.39 & 92.50 & 94.12 & 82.35 & 78.57 & 87.98 \\
& Fine-tuning \small(Multi) & 91.36 & 92.16 & 93.98 & 78.02 & 79.76 & 87.05  \\ 
& \textbf{Prompt-tuning} & 89.36 & 88.56 & 90.58 & 67.07 & 69.63 & 81.04 \\
& \textbf{\ours~(ours)} & 89.21 & 88.65 & 89.94 & 68.05 & 69.04 & 80.98 \\ \cmidrule{2-8}
& \textbf{$\rhd$ Pruning rates (\%)} & 36.7\% \small & 38.3\% \small & 55.0\% \small & 65.0\% \small & 65.0\% \small & 52.0\% \small \\
& \textbf{$\rhd$ Speed-up ($\times$)} & \color{red}$\times 1.29$ & \color{red}$\times 1.30$ & \color{red}$\times 1.43$ & \color{red}$\times 1.73$ & \color{red}$\times 1.72$ & \color{red}$\times 1.50$  \\
\bottomrule

\multirow{6}{*}{\makecell{RoBERTa}} & Fine-tuning \small(Single) & 94.34 & 94.56 & 94.61 & 85.13 & 86.90 & 91.10 \\
& Fine-tuning \small(Multi) & 94.65 & 94.70 & 94.64 & 86.85 & 72.02 & 88.57 \\
& \textbf{Prompt-tuning} & 93.84 & 92.72 & 91.87 & 67.23 & 66.06 & 82.34 \\
& \textbf{\ours~(ours)} & 92.81 & 91.50 & 90.79 & 67.89 & 69.04 & 82.40 \\ \cmidrule{2-8}
& \textbf{$\rhd$ Pruning rates (\%)} & 30.0\% \small & 36.7\% \small & 61.7\% \small & 58.3\% \small & 63.3\% \small  & 50.0\% \small \\ 
& \textbf{$\rhd$ Speed-up ($\times$)} & \color{red}$\times 1.18$ & \color{red}$\times 1.23$ & \color{red}$\times 1.25$ & \color{red}$\times 1.39$ & \color{red}$\times 1.53$  & \color{red}$\times 1.32$\\
\bottomrule 

\end{tabular}
}
\vspace{0.2cm}
\caption{
\textbf{Decomposition of Experts Results.} We report the accuracy of five datasets for \ours~and other baselines, such as two fine-tuning methods (upper bounds) and prompt-tuning. We train models using each method by three trials and report the averaged accuracy. Our method, \ours, achieves higher speed-ups (up to $\times$1.73) than other methods by leveraging high pruning rates (up to 65\%), while maintaining comparable performance to the existing prompt tuning method.}

\vspace{-0.5cm}
\label{table1}
\end{table*}

\noindent\textbf{Models and Training Methods.}\hspace{0.2cm}
We utilize two transformer variants\footnote{\url{https://huggingface.co/docs/transformers/}}, BERT \citep{devlin2018bert} and RoBERTa \citep{liu2019roberta}, for our experiments since they are some of the most popular basic language models.
We select a prompt tuning method \citep{lester2021power} as the backbone and baseline of our method since it is one of the basic PEFT methods.
We note that the typical LoRA \citep{hu2021lora} method is not appropriate for \ours, as it adds low-rank updates to the original weights at inference time, making parameter pruning difficult.
Additionally, we also include the results of basic fine-tuning and multi-task fine-tuning to show the upper-bound performance of each task.

\noindent\textbf{Implementation Details.}\hspace{0.2cm}
We evaluate \ours~and baselines on NVIDIA A5000 GPU.
We train models using $l=20$ prompt tokens during 100 epochs with early-stop conditions for all datasets.
In addition, we set the margin $\psi=1.0\%p$ for the validation accuracy score used in the Binary search algorithm.
We search $p \in [0.0, 0.05, ..., 0.95, 1.0]$ to find the best pruning rate in the Binary search algorithm.
We quantify the task relevance using only $n=20$ data instances of each training dataset, following \citet{yang2024mitigating}.
We specify the target modules for task expert localization as FFN networks since the experimental results for FFN networks outperform those of other modules (Described in Section~\ref{sec:module-specific}).
We use learning rates $\gamma_1 \in $ [1e-5, 5e-5] and $\gamma_2 \in $ [1e-4, 5e-3] for fine-tuning and prompt-tuning, respectively.
We predict output texts using a masked language model probing network for the prompts that include an instruction for solving a task, as shown in Table~\ref{table:templates}. 
Specifically, the text templates consist of an instruction and an input text.
For the prediction, we use the representation of the mask token (i.e., ``$<$\textit{mask}$>$").

\subsection{DoE is Efficient and Robust}
\label{exp:skill_results}
\begin{figure}[h]
\begin{minipage}[b]{1.0\linewidth}
  \centering
  \centerline{\includegraphics[width=0.95\linewidth]{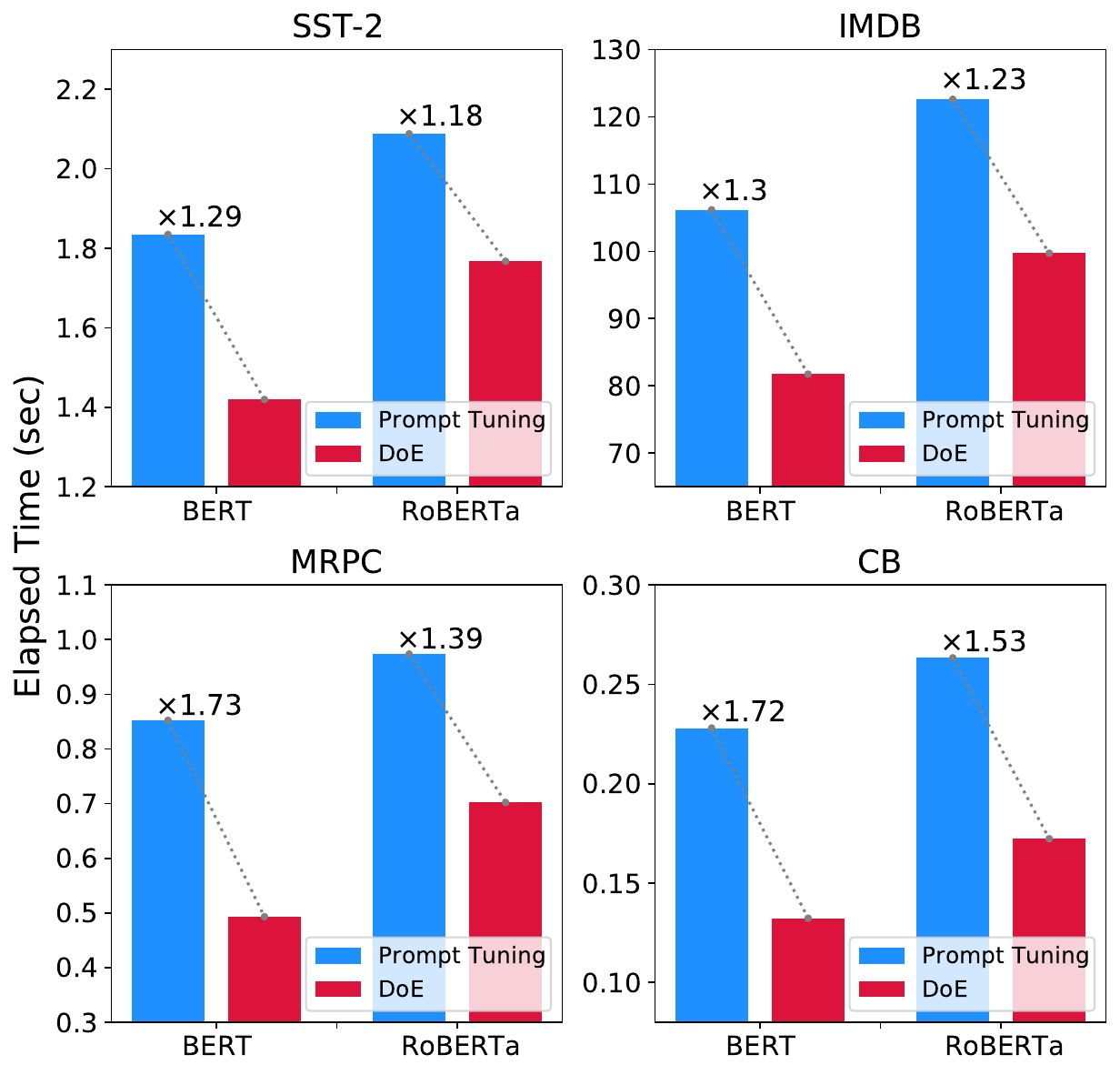}}
\end{minipage}
\caption{\textbf{Elapsed Time and Inference Speed-up.} We plot the elapsed time and inference speed-up for BERT-base and RoBERTa-base on the four datasets. The values above the bar plot are the speed-up ratios.}
\label{fig:inference_time}
\end{figure}

\noindent \textbf{Unplug-and-play Experiments.}\hspace{0.2cm}
We evaluate and report the task performance and efficiency of the baselines in Table~\ref{table1} for the five natural language understanding benchmarks.
We adopt BERT-base and RoBERTa-base to conduct the experiments.
The results show that our method performs comparably to the prompt tuning method, which utilizes the full model architecture, and even outperforms it in accuracy for some datasets using only a small number of task-localized parameters (e.g., at most 65\% pruning rates).
Surprisingly, the pruned networks show at most $\times 1.72$ speed up; thus, these results reveal that our method successfully localizes efficient task-relevant sub-networks from language models.
We describe the detailed elapsed time (sec) of \ours~models and prompt tuned model in Figure~\ref{fig:inference_time}.
These reveal that our method significantly outperforms the prompt tuning method in inference speed.

\begin{figure}[h]
\begin{minipage}[b]{1.0\linewidth}
  \centering
  \centerline{\includegraphics[width=1.0\columnwidth]{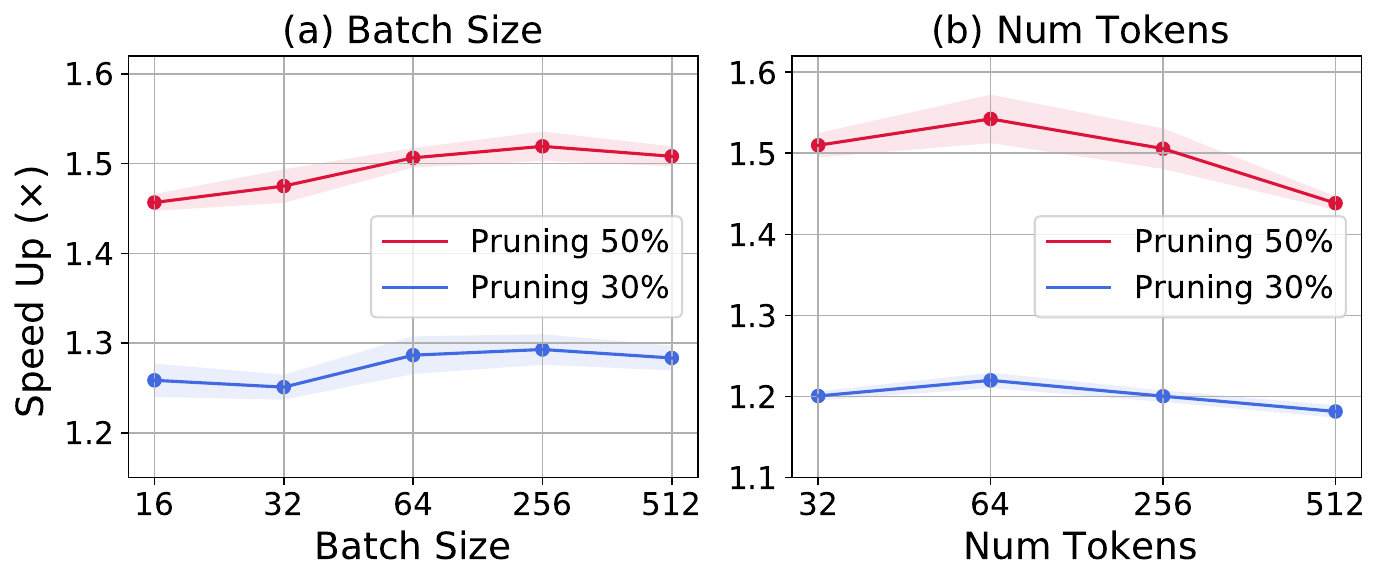}}
\end{minipage}
\caption{\textbf{Inference Speed Experiments for Varying Hyper-parameters (Batch size \& \# of Tokens).} We report the results for the pruning rates of 30\% and 50\% for BERT-base to ensure a fair comparison.} 
\label{fig:hyperparams}
\end{figure}

\noindent \textbf{Hyperparameter Experiments.} 
We conduct experiments to investigate the time efficiency of our framework for varying hyperparameters, as shown in Figure~\ref{fig:hyperparams}.
Specifically, we search varying batch sizes and the number of input tokens to demonstrate efficiency by comparing our framework to the prompt tuning method.
We sample random texts with $k$ tokens and input them into the BERT-base model for the experiments.
For the batch size experiments, we evaluate the inference speed with varying batch sizes $\beta \in [16, 32, 64, 256, 512]$, fixing the number of tokens at $64$.
For the number of tokens experiments, we measure the inference speed with multiple numbers of tokens $k \in [32, 64, 256, 512]$, using the batch size of $64$.
We conduct experiments on the fixed pruning rates $30\%$ and $50\%$ for the comparison.
The results reveal that our framework shows robust efficiency improvement in various hyperparameters, showing better speed in higher batch sizes and a widely used number of tokens.

\noindent \textbf{Experiments on various sizes of BERT.}
We conduct experiments to evaluate the scalability of our method to larger models, as shown in Table~\ref{table3}.
Specifically, we adopt BERT-large to show the scalability of our framework.
These results reveal the possibility that our method can be scaled to a larger model.

\begin{table}[h]
\centering
\resizebox{0.95\linewidth}{!}
{
\begin{tabular}{cccccc}
\toprule
\multirow{1}{*}{} & Method & SST-2 & IMDB \\ \midrule
\multirow{6}{*}{\textbf{\makecell{BERT-base}}} & Fine-tuning \small(Single) & 92.39 & 92.50 \\
& Fine-tuning \small(Multi) & 91.36 & 92.16 \\ 
& Prompt-tuning & 89.36 & 88.56 \\\cmidrule{2-4}
& \textbf{\ours} & 89.21 & 88.65 \\
& \textbf{$\rhd$ Pruning rates (\%)} & 36.7\% & 38.3\% \\
& \textbf{$\rhd$ Speed-up ($\times$)} & $\times$ 1.29 &  $\times$ 1.30 \\
\midrule
\multirow{6}{*}{\textbf{\makecell{BERT-large}}} & Fine-tuning \small(Single) & 92.77 & 93.57 \\ 
& Fine-tuning \small(Multi) & 92.73 & 93.83 \\ 
& Prompt-tuning & 91.28 & 89.96 \\\cmidrule{2-4}
& \textbf{\ours} & 92.19 & 89.35 \\
& \textbf{$\rhd$ Pruning rates (\%)} & 35\% & 30\% \\
& \textbf{$\rhd$ Speed-up ($\times$)} & $\times$ 1.34 & $\times$ 1.27 \\
\bottomrule
\end{tabular}
}
\caption{
\textbf{Experiments on various sizes of BERT.}
}
\vspace{-0.45cm}
\label{table3}
\end{table}

\subsection{Task Localization Methods Comparison}
\label{exp:task_local_method}
We compare our method to other task neuron selection methods: \textit{Activation}, \textit{Gradient}, and \textit{Random}, to justify the excellence of the attribution method (Section~\ref{exp:task_local_method}).
\noindent \textbf{\textit{Activation}} identifies task neurons by using the activated values of neurons. We compute the sum of activated values of each neuron when inputting a dataset.
\noindent \textbf{\textit{Gradient}} selects task neurons by using the gradient values of each neuron. We compute the sum of the gradient values of each neuron when predicting the label of the utilized dataset.
\noindent \textbf{\textit{Random}} randomly selects task neurons.

\begin{figure}[h]
\begin{minipage}[b]{1.0\linewidth}
  \centering
  \centerline{\includegraphics[width=0.7\linewidth]{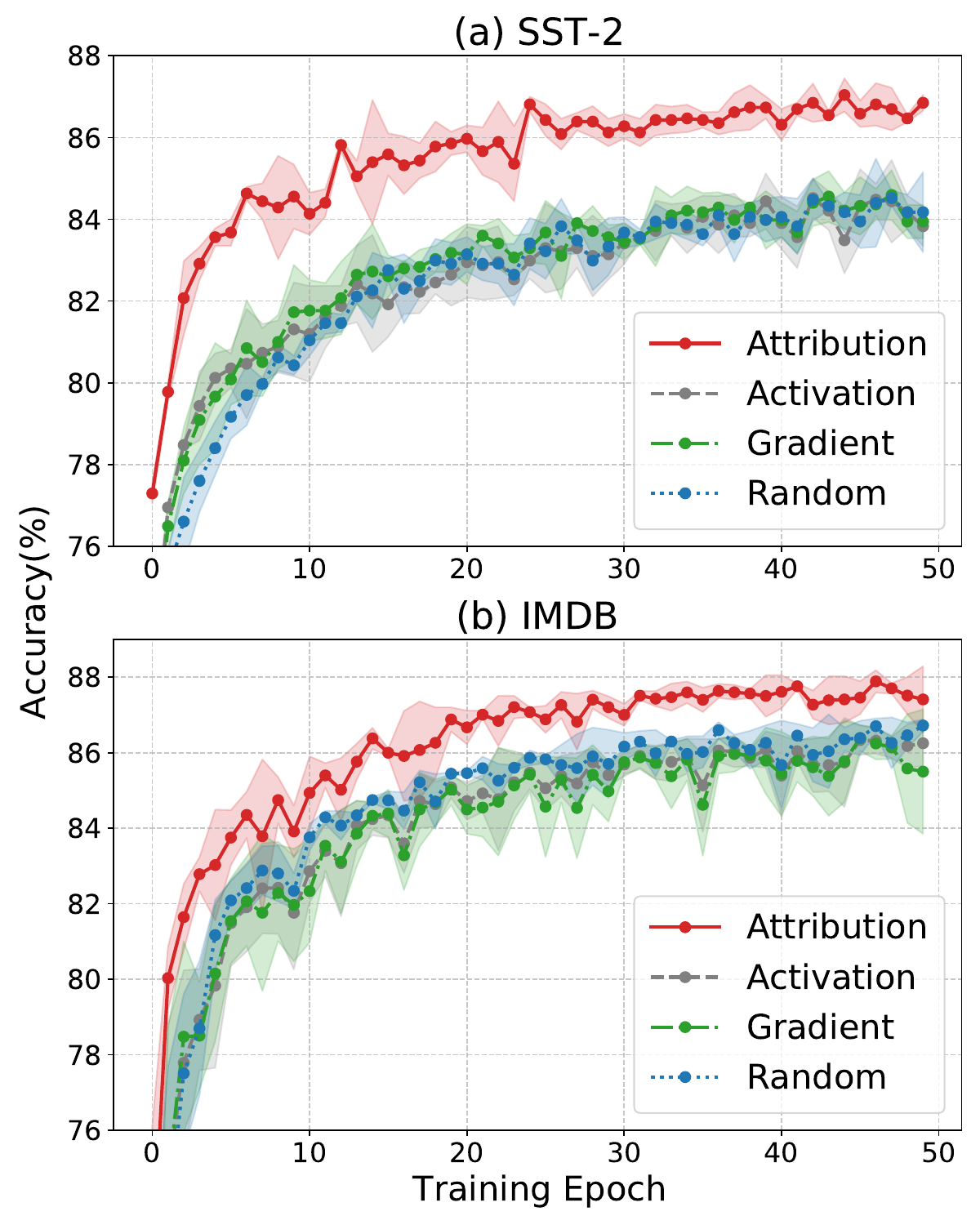}}
 \vspace{-0.35cm}
\end{minipage}
\caption{\textbf{Task Localization Methods Experiments.} We plot the mean accuracy (± one standard deviation for three trials) of BERT-base for SST-2 and IMDB.}
\label{fig:fig_methods}
\vspace{-0.35cm}
\end{figure}

We demonstrate the task localization performance of those baselines to justify our selection of the task localization method, \textit{Attribution}, as shown in Figure~\ref{fig:fig_methods}.
Specifically, we select the BERT-base model for the SST-2 and IMDB datasets for the experiments.
We prune the top 50\% of task neurons by using those methods for a fair comparison.
Surprisingly, even \textit{Activation} and \textit{Gradient} do not show competitive performance compared to \textit{Random}.
However, \ours~(i.e., \textit{Attribution}) outperforms other methods in the two datasets, revealing the excellence of our method.

\begin{table*}[ht] 
\centering
\resizebox{1.0\textwidth}{!}
{
\begin{tabular}{ccc} 
\toprule
\textbf{Dataset} & \textbf{Template} & \textbf{Label} \\
\midrule
SST-2   & Text: \textcolor{blue}{\{text\}}. The sentiment of the text is \textcolor{gray}{$<$mask$>$}. & positive / negative \\
IMDB    & Text: \textcolor{blue}{\{text\}}. The sentiment of the text is \textcolor{gray}{$<$mask$>$}. & positive / negative \\
AGNews  & Text: \textcolor{blue}{\{text\}}. The topic of the text is \textcolor{gray}{$<$mask$>$}. & World / Sports / Business / Science \\
MRPC    & Text1: \textcolor{blue}{\{text1\}}. Text2: \textcolor{blue}{\{text2\}}. The two texts are \textcolor{gray}{$<$mask$>$}. & different / equivalent \\
CB      & Premise: \textcolor{blue}{\{premise\}}. Hypothesis: \textcolor{blue}{\{hypothesis\}}. \par The premise and hypothesis have a relationship of \textcolor{gray}{$<$mask$>$}. & implication / contradiction / neutrality \\
\bottomrule
\end{tabular}
}
\caption{\textbf{Instruction Templates and Labels for Each Dataset.} The templates show how text, premise, hypothesis, and mask tokens are used in each dataset.}
\vspace{-0.5cm}
\label{table:templates}
\end{table*}

\subsection{Module-specific Task-localization}
\label{sec:module-specific}
\begin{figure}[H]
\begin{minipage}[b]{1.0\linewidth}
  \centering
  \centerline{\includegraphics[width=0.95\linewidth]{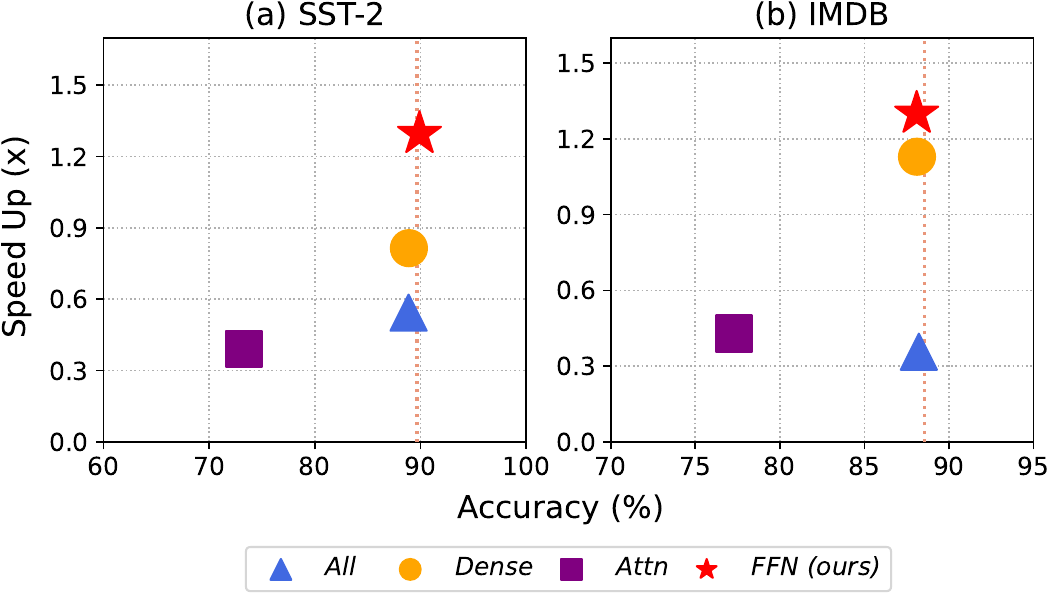}}
 \vspace{-0.2cm}
\end{minipage}
\caption{\textbf{Module-specific Task-localization Results.} We plot the relationship between the accuracy and speed up of each module, where each diagram corresponds to the categorized module. The red-dotted line is the baseline performance (Prompt-tuning).}
\label{fig:fig_module}
\end{figure}

Transformer variants have various types of modules (e.g., attention modules and FFNs).
Therefore, we examine which module is suitable for maintaining the knowledge of a language model (BERT-base).
Specifically, we categorize layers into four segments: (1) All layers (\textit{All}), (2) Attention layers (\textit{Attn}), (3) Dense layers (\textit{Dense}), and (4) Feed-forward layers (\textit{FFN}).
\textit{All} includes all linear layers.
\textit{Attn} consists of Query (Q), Key (K), Value (V), and Output (O) layers of the attention module.
\textit{Dense} contains $O$ of the attention module and two FFNs.
\textit{FFN} includes two FFNs.
We prune the top 30\% of task neurons from each module for a fair comparison.
Figure~\ref{fig:fig_module} represents the results of our method for various types of modules.
From these experiments, we reveal that FFNs are suitable for maintaining the knowledge of a language model in our prompt tuning setting.
Adopting the \textit{FFN} module yields faster performance than other modules without compromising accuracy.

\subsection{Ablation Study on Knowledge Condensation}
We conduct the ablation study to justify the necessity of Knowledge Condensation, as described in \textit{The Knowledge Condensation Step} (Section~\ref{task_neuron_loc_prompt_tuning}).
Specifically, we eliminate neurons the same as the results of the \ours~in the ablation study for a fair comparison, and do not conduct the knowledge condensation step.
As a result, if we do not use Knowledge Condensation, the performance for SST-2 and IMDB significantly degrades.
These results reveal the justification for adopting the knowledge condensation step.

\begin{table}[h]
\centering
\resizebox{1.0\linewidth}{!}
{
\begin{tabular}{cp{4.7cm}ccc}
\toprule
&  & SST-2 & IMDB & \\ \midrule
& \textbf{\ours} & 89.21 & 88.65 & \\
& \textbf{w/o Knowledge Condense} & 0.0 & 48.34 & \\ \midrule
& \textbf{$\rhd$ Pruning rates (\%)} & 36.7\% & 38.3\% & \\
\bottomrule
\end{tabular}
}
\vspace{0.4cm}
\caption{
\textbf{Ablation Studies on Knowledge Condensation.}
}
\label{table4}
\end{table}

\subsection{Prior Knowledge Aligning is Important}
\begin{figure}[h]
\begin{minipage}[b]{0.82\linewidth}
  \centering
  \centerline{\includegraphics[width=0.8\linewidth]{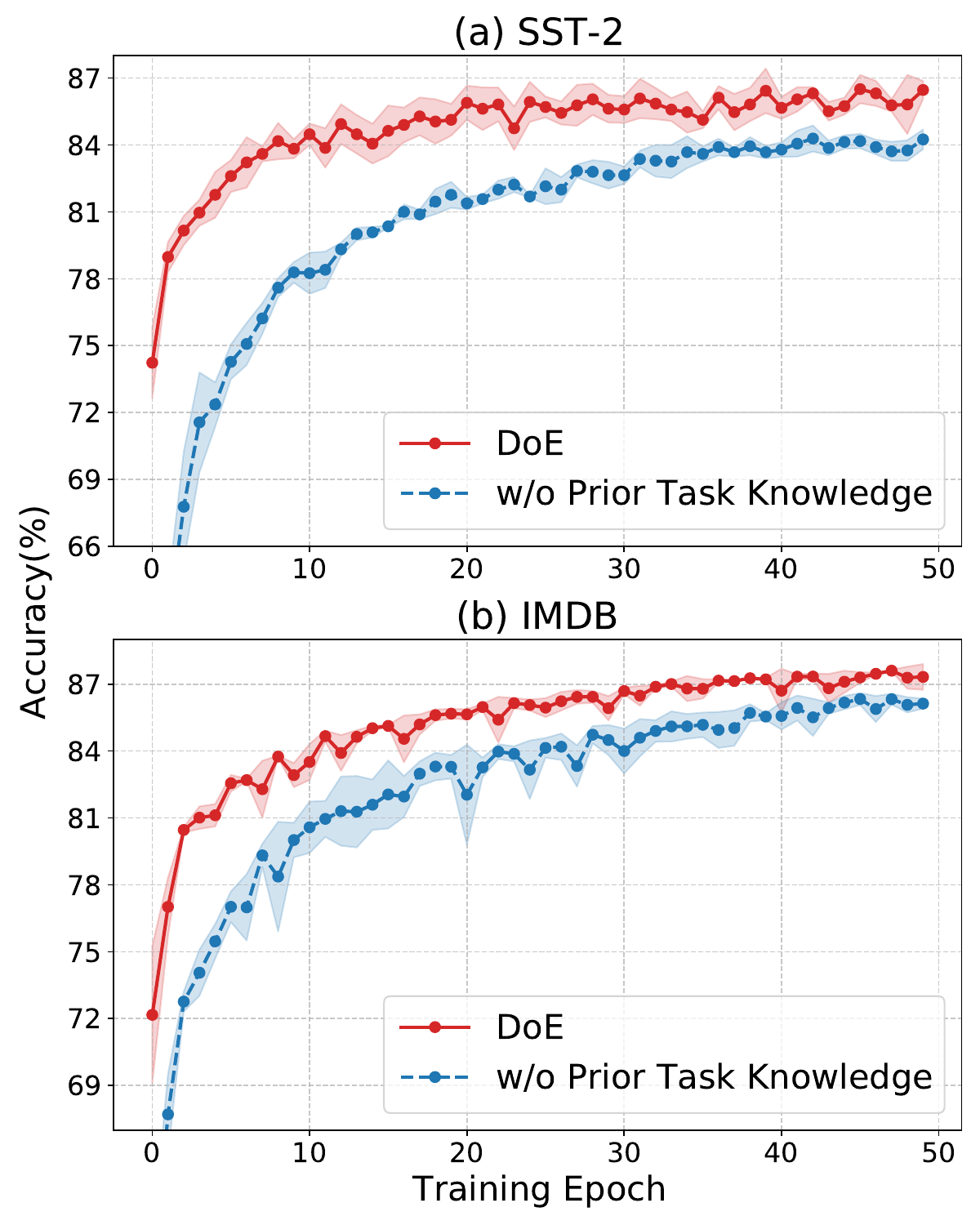}}
 \vspace{-0.2cm}
\end{minipage}
\caption{\textbf{Prior Knowledge Injection is Important.} We plot the mean accuracy (± one standard deviation for three trials) of BERT-base for SST-2 and IMDB.}
\label{fig:fig_prior}
\vspace{-0.3cm}
\end{figure}

To accurately quantify task relevance, this section highlights the role of aligning prior task knowledge using prompt tuning, as described in \textit{The Knowledge Quantification Step} (Section~\ref{task_neuron_loc_prompt_tuning}) and illustrated in Figure~\ref{fig:fig_prior}.
We prune the top 50\% of task neurons by using two methods (i.e., \textit{\ours}~and \textit{w/o Prior Task Knowledge}) for a fair comparison.
The experimental results reveal that prior task knowledge alignment has a crucial role in quantifying the task relevance.
Excluding the Prior Knowledge Alignment results in slower convergence and lower task accuracy than our method.

\section{Conclusion}

In this study, we proposed Decomposition of Experts (\ours), a novel framework that optimizes pre-trained language models by identifying and activating task-specific experts.
To the best of our knowledge, our method is the first to propose a novel usage paradigm for language models that temporarily deactivates irrelevant experts (i.e., neurons) and restores the full model to await subsequent tasks.
\ours~leverages an unplug-and-play strategy to enhance inference efficiency while maintaining task performance. Experimental results show that \ours~achieves up to a 1.73x speed-up with a 65\% parameter pruning rate across diverse tasks. Comparative evaluations and ablation studies validate the effectiveness of \ours~and its components. The scalability of \ours~makes it a practical and adaptable solution for transformer-based architectures.

\newpage

\begin{acks}
This work was partly supported by Institute of Information \& communications Technology Planning \& Evaluation (IITP) grant funded by the Korea government (MSIT) [No.RS-2022-II220184, Development and Study of AI Technologies to Inexpensively Conform to Evolving Policy on Ethics \& No.RS-2021-II211343, Artificial Intelligence Graduate School Program (Seoul National University) \& No.RS-2021-II212068, Artificial Intelligence Innovation Hub (Artificial Intelligence Institute, Seoul National University)].
K. Jung is with ASRI, Seoul National University, Korea.
The Institute of Engineering Research at Seoul National University provided research facilities for this work.
\end{acks}

\section*{GenAI Usage Disclosure}
Generative AI was not used at any stage of the research, except for minor corrections related to spelling or grammar in the writing.

\section*{Appendix}
\appendix

\section{Module-specific Task-localization Details}
We conduct experiments for various modules in the transformer architecture, as described in Section~\ref{sec:module-specific}.
We implement algorithms for pruning networks following our \ours~framework to satisfy the dimension of adjacent networks after pruning.
For example, suppose that we prune the Key network, $W^{K} \in \mathbb{R}^{d \times d}$, and get a pruned weight, $\tilde{W}^{K} \in \mathbb{R}^{d \times d'}$, in the Attention module. Then, we also prune the Query network, $W^{K} \in \mathbb{R}^{d \times d}$, since it is an adjacent network.
Therefore, if we get $v_Q \in \mathbb{R}^{d}$ and $v_K \in \mathbb{R}^{d'}$ from the Query and Key networks after pruning, we recover the pruned parameters of $v_K \in \mathbb{R}^{d'}$ by inserting zero values to ensure the identical dimension.
We also implement other layers in the same way.

\bibliographystyle{ACM-Reference-Format}
\bibliography{sample-base}

\end{document}